\pdfoutput=1

\documentclass[11pt]{article}

\usepackage{naacl2021}
\usepackage{todonotes}
\usepackage{times}
\usepackage{latexsym}

\usepackage{listings}
\usepackage{xcolor}

\colorlet{punct}{red!60!black}
\definecolor{background}{HTML}{EEEEEE}
\definecolor{delim}{RGB}{20,105,176}
\colorlet{numb}{magenta!60!black}

\lstdefinelanguage{json}{
    basicstyle=\normalfont\ttfamily,
    numbers=left,
    numberstyle=\scriptsize,
    stepnumber=1,
    numbersep=8pt,
    showstringspaces=false,
    breaklines=true,
    frame=lines,
    backgroundcolor=\color{background},
    literate=
     *{0}{{{\color{numb}0}}}{1}
      {1}{{{\color{numb}1}}}{1}
      {2}{{{\color{numb}2}}}{1}
      {3}{{{\color{numb}3}}}{1}
      {4}{{{\color{numb}4}}}{1}
      {5}{{{\color{numb}5}}}{1}
      {6}{{{\color{numb}6}}}{1}
      {7}{{{\color{numb}7}}}{1}
      {8}{{{\color{numb}8}}}{1}
      {9}{{{\color{numb}9}}}{1}
      {:}{{{\color{punct}{:}}}}{1}
      {,}{{{\color{punct}{,}}}}{1}
      {\{}{{{\color{delim}{\{}}}}{1}
      {\}}{{{\color{delim}{\}}}}}{1}
      {[}{{{\color{delim}{[}}}}{1}
      {]}{{{\color{delim}{]}}}}{1},
}
\usepackage[T1]{fontenc}

\usepackage[utf8]{inputenc}

\usepackage{microtype}

\usepackage{graphicx}

\title{From Judgement's Premises Towards Key Points}

\author{Oren Sultan* \\ \small{The  Technical University of Munich}, \\ \small{The Hebrew University of Jerusalem }\\ 
\small {\texttt{oren.sultan@mail.huji.ac.il}} \\\And
  Rayen Dhahri* \\
  \small{The Technical University of Munich} \\
  \small {\texttt{rayen.dhahri@tum.de}} \\\AND
  Yauheni Mardan* \\
  \small{The  Technical University of Munich} \\
  \small {\texttt{yauheni.mardan@tum.de}} \\\And 
  Tobias Eder \& Georg Groh \\
  \small {The  Technical University of Munich} \\ 
  \small {\texttt{\{tobias.eder, grohg\}@in.tum.de}} \\
  }

\begin{document}
\maketitle

\footnote{* These authors contributed equally}

\begin{abstract}
    
\textbf{Key Point Analysis(KPA)} is a relatively new task in NLP that combines summarization and classification by extracting argumentative key points (KPs) for a topic from a collection of texts and categorizing their closeness to the different arguments.

\textbf{In our work}, we focus on the \textbf{legal domain} and develop methods that identify and extract KPs from \textbf{premises} derived from texts of \textbf{judgments}. 
The first method is an adaptation to an existing state-of-the-art method, and the two others are new methods that we developed from scratch. We present our methods and examples of their outputs, as well a comparison between them. The full evaluation of our results is done in the \textbf{matching task} -- match between the generated KPs to arguments (premises).

\end{abstract}

\section{Introduction}

\textbf{Key Point Analysis (KPA)} was introduced in \citet{bar-haim-etal-2020-arguments}, \citet{bar-haim-etal-2020-quantitative} as a new challenging NLP task with close relation to Computational Argumentation, Opinion Analysis, and Summarization. Given a collection of relatively short, opinionated texts focused on a topic of interest, the goal of KPA is to produce a concise list of the most prominent \textbf{Key Points (KPs)}, but also to quantify their \textbf{prevalence}. KPA can be used to gain better insights from public opinions expressed through social media, surveys, etc.

\textbf{In our work}, we focus on the \textbf{legal domain} -- given a collection of \textbf{premises} from \textbf{judgments}, our goal is to extract the most prominent KPs from the premises. We explored three different methods for the KP extraction task. We show examples of outputs of our methods and a comparison between them. Our contributions are:
\begin{itemize}
    \item We present three new methods for KP extraction in a new domain of legal dataset -- extracting KPs from premises of judgment texts.
    \item We present a comparison between the outputs of our methods. Full evaluation is included in the work of the matching task.
    \item We release data and code, including our dataset, as well as the outputs of our different methods on the dataset.  \footnote{https://gitlab.lrz.de/lab-courses/nlp-lab-ss2022/team-list-2-tobias}
\end{itemize}

We hope this work will inspire the development of new methods and diverse datasets to be applied to this new task.

\section{KP Extraction Task}

Given a collection of arguments towards a certain topic, the goal is to generate KP-based summary. KPs should be concise, non-redundant, and capture the most important points of a topic. Ideally, they should summarize the input data at the appropriate granularity -- general enough to match a significant portion of the arguments, yet informative enough to make a useful summary. Then, in the \textbf{matching task}, the goal is to compute the confidence score between arguments to the extracted KPs. 

\textbf{In our work}, the arguments are premises from different texts of judgments, and we apply this task on them. 

\section{Dataset}

\begin{figure*}[t]
\begin{centering}
\includegraphics[scale=.22]{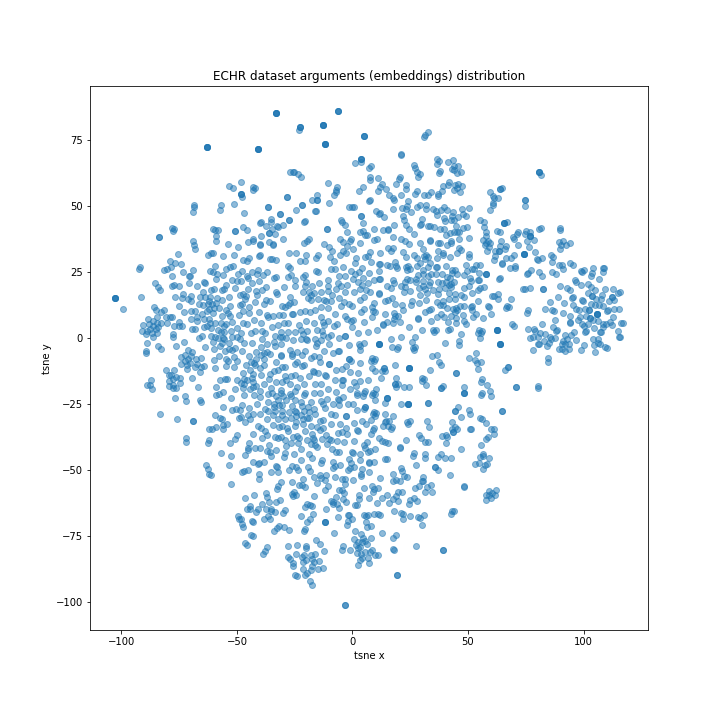}
\includegraphics[scale=.22]{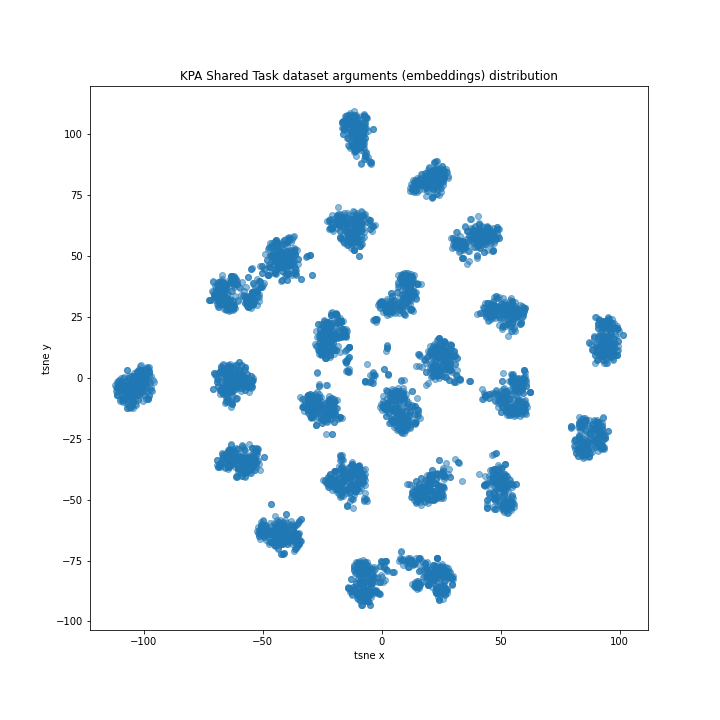}
\par\end{centering}
\caption{TSNE plots which illustrate the difference between embeddings distributions of our dataset (\textbf{left}) and ArgKP (\textbf{right}). Our argument's embeddings are denser, while in the argKP dataset, they are highly-clustered. Hence, in our case, it will be more challenging to find clusters of arguments and generate suitable KPs.}
\label{fig:datasets}
\end{figure*}

We use the \textbf{European Court of Human Rights (ECHR)} dataset to extract KPs in all of our methods. It consists of 42 human-annotated \textbf{judgements}. The corpus is annotated in terms of \textbf{premises} and \textbf{conclusion}. Overall, 1951 premises and 743 conclusions. We consider the \textbf{premises} as \textbf{arguments}, and extract the KPs from them.

\textbf{Example of an argument}: 

\emph{``The Commission considers that this indicates an issue falling within the scope of freedom of expression.''}.

In the work of \citet{bar-haim-etal-2020-arguments}, they developed a large-scale labeled dataset for the task of matching arguments to key points. The
dataset, termed \textbf{ArgKP}, contains about 24K (argument, key point) pairs, for 28 controversial topics. Each of the pairs is labeled as matching/nonmatching. 

\citet{bar-haim-etal-2020-arguments} experimented with \textbf{BERT} \citet{devlin-etal-2019-bert} as a supervised model for argument matching, which they trained on the \textbf{ArgKP} dataset. Later, \citet{bar-haim-etal-2020-quantitative} experimented with more transformers-based LMs.

We compare the embeddings of arguments in the \textbf{ArgKP} 
to ours, we find our argument's embeddings are denser, which makes it harder to find clusters of arguments, and get a good result in the end task (in comparison to the results on \textbf{ArgKP}). See Figure~\ref{fig:datasets} for the different between argument's embeddings over the datasets.

\label{sec:dataset}

\section{Method I : KP Candidate Extraction and Selection Using IBM Debater}
We use the KPA method of \citet{bar-haim-etal-2020-quantitative} which extracts KPs in two steps. 
\begin{itemize}
        \item \textbf{Candidate Extraction -- find KP candidates from the arguments}. 
        
        When extracting candidates, this method assumes that KP can be found in the given arguments. To get concise and high quality candidates, we choose only single sentences, and filter sentences with less than \textbf{4} tokens or more than \textbf{36} tokens. 
        
        To ensure quality of candidates, the method uses IBM-ArgQ-Rank-30kArgs dataset which consists of around 30k arguments annotated for point-wise quality to train an argument quality ranking model. The method uses the model to include only high quality candidates. 
        
        Lastly, in order to keep the key points self-contained, sentences starting with pronouns are filtered.

        \item \textbf{KP Selection -- the most prominent candidates are selected as KP}. Now, we use a matching model (between KPs to arguments) as described in \citet{bar-haim-etal-2020-quantitative}(Section 2). So, KPs which are not matched with arguments will not be selected. We chose a mapping threshold of \textbf{0.9} (a higher threshold leads to higher precision and a lower coverage). We fine-tuned the threshold manually, and decided about 0.9 as we care more about precision rather than to cover all the data. When running the matching model with this threshold on all the judgment's texts together, it achieves 0.45 sentence coverage and 0.67 arguments coverage.
        
        We match a sentence to only \textbf{one} KP (the one with the highest matching score). 
    \end{itemize}
   In the output of the method, one argument is broken into multiple sentences (each possibly connected to a different KP). Hence, we concatenated the sentences back to the original argument and chose the KP with the maximum score. We propose two pipelines:
    \begin{itemize}
        \item Create KPs out of all texts.
        \item \textbf{Create KPs for every text separately and union the KPs}
    \end{itemize}
    Since judgements are independent, it makes more sense to go with the second pipeline. Indeed, it creates higher-quality KPs. But, on the downside, since we run the method on every text separately and some of them can be very short, it is possible that no KPs will be generated, for some texts.

\section{Method II : Clustering and Summarization}

This method consists of two steps: arguments clusterization and summarization of clustered arguments. First, semantically similar arguments are grouped together, and then we create a summary for each resulting group of arguments. We consider every summary as a new key point.

\subsection{Arguments Clusterization}
\begin{figure}[t]
\begin{centering}
\includegraphics[scale=.3]{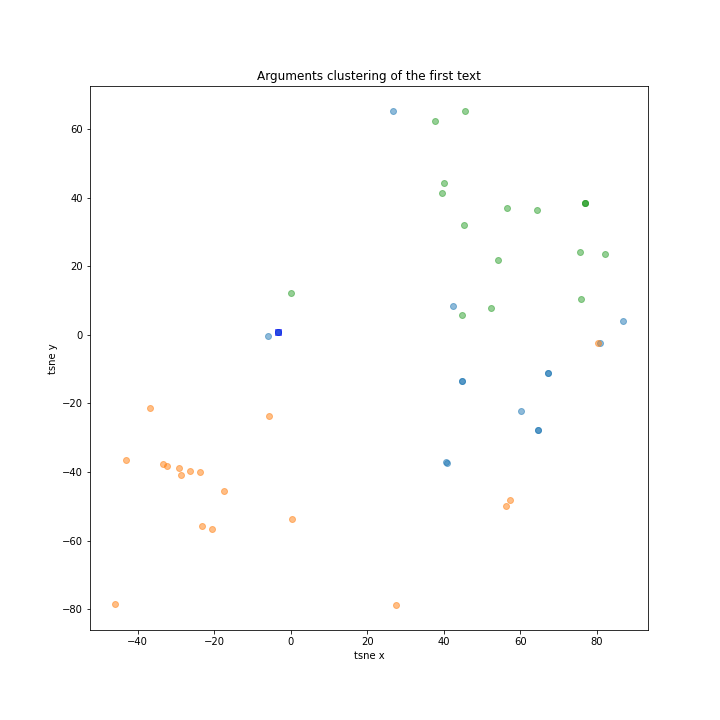}
\par\end{centering}
\caption{TSNE plot with the embeddings of arguments in the first text. The blue square here is the argument shown in Section \ref{sec:dataset}. As we can see, the argument's embeddings can be divided into different, highly separated clusters.}
\label{fig:clustering_example}
\end{figure}

First, we use \textbf{Legal-BERT} \cite{https://doi.org/10.48550/arxiv.2010.02559} to encode arguments to embeddings. Then, we cluster embeddings of the arguments for each text separately using a clusterization algorithm. We consider two different algorithms: \textbf{Hierarchical Agglomerative Clustering} \cite{Zepeda-Mendoza2013} and \textbf{HDBSCAN} \cite{https://doi.org/10.48550/arXiv.1911.02282}. We manually evaluated the resulting clusters and found that \textbf{HDBSCAN} achieves better results. 
At the end of this step, we have several clusters of arguments for each text. See Figure~\ref{fig:clustering_example} for an example of arguments clusterization on the first text.

\subsection{Summarization of Clusters}
Now, we apply summarization on each cluster of arguments. During summarization, all arguments within one cluster are concatenated and passed to a summarization model as one text. We suggest two approaches:

\begin{itemize}
\item \textbf{Extractive}:
We consider the following extractive summarization models: \textbf{LexRank} \cite{https://doi.org/10.48550/arXiv.1109.2128}, \textbf{LSA}, \textbf{Luhn} \cite{http://dx.doi.org/10.1147/rd.22.0159}, and \textbf{KL-Sum}. 

In extractive models, the resulting KP is one of the original arguments.

\item \textbf{Abstractive}:
We consider the following abstractive models: 
\textbf{BART}  \cite{https://doi.org/10.48550/arXiv.1910.13461}, \textbf{Pegasus}  \cite{https://doi.org/10.48550/arXiv.1912.08777}, and \textbf{Legal-Pegasus}\footnote{\url{https://huggingface.co/nsi319/legal-pegasus}} (fine-tuned on legal dataset version of Pegasus). 
We noticed that the resulting summarization of Legal-Pegasus includes critical legal terms. On the other hand, models which are not trained on a legal dataset omitted critical legal terms very often. 

In abstractive models, the summarization is a new sentence, that does not appear in the text.
\end{itemize}

Since \textbf{HDBSCAN} and \textbf{Legal-Pegasus} perform better for the clusterization and summarization steps respectively, their combination is used in the final version of this KP extraction method.

\section{Method III: PageRank and Clustering}

In this method, we use a fast version of PageRank algorithm \footnote{\url{https://github.com/asajadi/fast-pagerank}} to extract the KPs. The KPs are arguments that are already existing in the text.

\subsection{PageRank}
PageRank \cite{Page1998PageRank} is an algorithm mainly known for it being used to rank the web pages by relevancy to display once using a search engine. PageRank computes and counts the number and quality of links to a page to determine a rough estimation of the importance of the website according to the user's search query.

\textbf{In our work}, PageRank is used to rank the arguments as potential KP candidates by their relevancy according to a similarity score between the embeddings of the arguments.

For the task of KP extraction, \textbf{SMatchToPR} \cite{https://doi.org/10.48550/arxiv.2109.15086} was the first method  to introduce PageRank in the KP shared task \cite{friedman-etal-2021-overview}.
The method consists of selecting the arguments with high-quality scores as KP candidates. In this method, the KP candidates were already available in the dataset and clustered by a labeled topic. Each KP candidate was encoded and given to the \textit{argument quality score} function obtained from IBM Debater to select which KP candidates can be an actual KP.  
Then, the Rouge score \cite{lin-2004-rouge} is computed between the ground truth KP and the 10 best-ranked sentences between all arguments, to select the final KP (that surpasses a threshold) \cite{https://doi.org/10.48550/arxiv.2109.15086}.

Note that for the ArgKP dataset (Section \ref{sec:dataset}), arguments can be composed of multiple sentences, resulting in some KP being a sentence of an argument.

\subsection{Method's Challenges}\label{methch}
As we don't have the topics of arguments in our legal dataset (Section \ref{sec:dataset}), we cannot apply methods like SMatchToPR directly.

To overcome this issue, we first tried to generate the topics by clustering the arguments and applying an abstractive summarization method using Pegasus. When applying SMatchToPR using Pegasus, we face two issues. Pegasus gave bad quality topics, and IBM's \textit{Argument quality score} function gives relatively high scores even though the argument and the topic do not match.

\subsection{Our Approach}
\begin{figure}[ht]
\begin{centering}
\includegraphics[scale=.55]{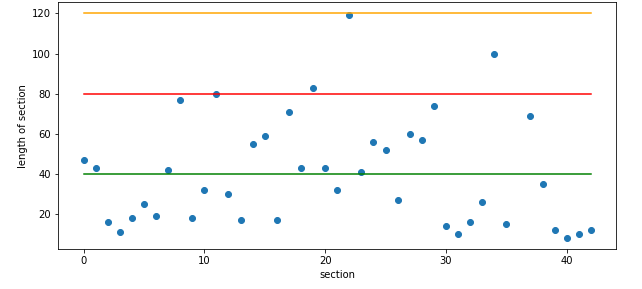}
\par\end{centering}
\caption{Number of arguments per text. The lines are used to set the number of KP to extract based on the distribution. Below the green line (40 arguments), we extract 3 KPs; between the green and red line (80 arguments), 6 KP; between the red and orange line (120 arguments), 9 KPs; and above 120 arguments, 12 KPs.}
\label{fig:numarg}
\end{figure}

To tackle the issues mentioned above, we chose to use a different similarity score between the argument's embeddings and to perform PageRank in two different ways.
\begin{itemize}
    \item \textbf{Quantitative} based: 
    we concatenate the arguments of each judgment's text and perform \textbf{PageRank} by first encoding the arguments to embeddings using \textbf{LegalBERT}. Then, we calculate the cosine similarity matrix between them. The ones with the highest number of comparisons that pass a \textbf{minimum threshold} (\textbf{0.8}) are KP candidates.
    Then, from these KP candidates, we choose \textbf{N} KPs with cosine similarity score below a \textbf{maximum threshold} of \textbf{0.4}, in order to avoid covering semantically similar KPs. 
    \textbf{N}, \textbf{maximum threshold} are hyperparameters, manually tuned, based on the number of arguments. We use the number of arguments per text to set the number of KPs to extract. See Figure~\ref{fig:numarg} for the distribution of the number of arguments per text, with respect to the number of KPs to be extracted.

    \item \textbf{Clustering} based:
    First, we cluster the arguments of each judgment's text using \textbf{HDBSCAN} algorithm. Then, we repeat the same steps of the first approach from the encoding to the hyperparameter's selection. Here, we tune the \textbf{maximum threshold} by using the \textbf{distance between cluster's centroid}, and we set \textbf{N} to be the number of clusters per judgment text.
        
\end{itemize}

The clustering based approach usually extracts less KPs than the quantitative approach, but still captures the main ideas of the text. 

\section{Other Methods}

\begin{itemize}
    \item \textbf{Prepare Potential KPs}:
    This method gets as an \textbf{input} the argument, topic and the score between them. Then, only \textbf{high quality sentences} remain, by taking sentences with scores above a threshold and sentences which are not too short and not too long. Then, \textbf{cross sentences} are applied --  get \textbf{SBERT} \cite{reimers-2019-sentence-bert} similarity of all pairs of sentences and create a new data frame containing  (Sentence1, Sentence2, Topic, SBERTScore, score1, score2). We take sentence1 above similarity threshold, and remove \textbf{redundant sentences}. The \textbf{output} are sentences as potential KPs.
    
    We didn't use this method, since it creates \textbf{low-quality KPs}, compare to other methods. Note that it is possible to get better results with a better topic-modeling generation method. We used the topic-modeling mentioned in \ref{methch}.
    
    \item \textbf{Enigma}:
    \citet{https://doi.org/10.48550/arXiv.2110.12370} presents a KP extraction method based on arguments rephrasing (using \textbf{Pegasus}) and filtering. This method needs actual KPs during the training and testing steps. As our dataset does not contain actual KPs, we didn't use this method.
\end{itemize}

\section{Methods Output Comparison}

Here is the premise from Section \ref{sec:dataset}:
\emph{``The Commission considers that this indicates an issue falling within the scope of freedom of expression.''}.

\textbf{The KP that each method generated}:

\textbf{Method I:}
\emph{``Everyone has the right to the freedom of expression''}

\textbf{Method II:}
\emph{``The Commission finds no evidence in the case to substantiate this complaint''}

\textbf{Method III:}
\emph{``The Commission finds that the applicant was deprived of his liberty after conviction by a competent court within the meaning of Article 5 para''}

As we can see, \textbf{Method I} generated a short KP which captures the main idea of the argument. This KP covers 12 different arguments.

\textbf{Method II} generated a KP that covers 16 different arguments. The KP conveys the main sense of the corresponding argument. However, some details are missed, like ``freedom of expression''.

\textbf{Method III} succeeded to capture the main idea of the argument. However, it is longer than the argument itself. However, this KP covers the highest number of arguments -- 45 different arguments.

\section{Conclusions and Future Work}

In this work, we tackled the new task of KP extraction, on a new dataset from the \textbf{legal domain}. We developed three different methods to extract KPs from premises (arguments) of judgment's texts. 
We demonstrated the outputs of the methods on one argument and compared between them. Notice that, to be able to compare the methods by metrics, they should be evaluated as part of the \textbf{matching task}.

Some important remarks about the methods:
\begin{itemize}
\item In \textbf{methods I and III} we create KPs which have already appeared in the arguments, while in \textbf{method II} we are able to create KPs which are new sentences.
\item In \textbf{method I}, there is a tradeoff between \textbf{coverage} (how many arguments are linked to a KP out of all arguments) and the \textbf{precision} (how many KPs are correctly matched to arguments) of the KPs. In our case, we prioritize higher precision and less coverage.
\item \textbf{Method II} is more \textbf{flexible}. It enables us to choose different clustering and summarization algorithms.
\item \textbf{Method III} allows us to determine the number of KP to extract and their granularity \textbf{in advance}, thanks to the filtering pipeline.
\end{itemize}

In the future, we plan to get the evaluation from the matching task, and then to decide which method works the best for this dataset and fine-tune on it.

We release our code and the dataset, as well as the full outputs of our different methods. 

\section*{Acknowledgements}
This research project is done as part of a NLP Lab course at the department of Informatics, The Technical University of Munich.

We would like to thank M.Sc., M.A Tobias Eder for his assistance and support throughout our work on this project. We also thank the other students who participated in the course for listening and discussing our ideas.

\bibliography{custom}
\bibliographystyle{acl_natbib}




\end{document}